\documentclass[11pt]{article}

\usepackage[pdftex]{graphicx}
\usepackage{hyperref}
\usepackage{amssymb}
\usepackage{amsmath}
\usepackage{bm}
\usepackage{mathtools}
\usepackage{booktabs}
\usepackage{algorithm}
\usepackage[noend]{algorithmic}
\usepackage[sort&compress,numbers]{natbib}
\usepackage{authblk}

\title{Discovering Effective Policies for Land-Use Planning\\with Neuroevolution}

\title{Discovering Effective Policies for Land-Use Planning\\with Neuroevolution}
\author[1,2]{Daniel Young}
\author[1]{Olivier Francon}
\author[1]{Elliot Meyerson}
\author[3]{Clemens Schwingshackl}
\author[4]{Jacob Bieker}
\author[5]{Hugo Cunha}
\author[1]{Babak Hodjat}
\author[1,2]{Risto Miikkulainen}

\affil[1]{Cognizant AI Labs, San Francisco, USA}
\affil[2]{The University of Texas at Austin, USA}
\affil[3]{Ludwig-Maximilians-Universit\"{a}t, Munich, Germany}
\affil[4]{Open Climate Fix, London, United Kingdom}
\affil[5]{Cognizant Technology Solutions, Brussels, Belgium}

\small\normalsize
\date{}
\begin{document}
\maketitle

\begin{abstract}
How areas of land are allocated for different uses, such as forests, 
urban areas, and agriculture, has a large effect on the terrestrial carbon balance, and
therefore climate change. Based on available historical data on land-use changes
and a simulation of the associated carbon emissions and removals, a surrogate model can be learned that makes it possible to evaluate the different options available to decision-makers efficiently. An 
evolutionary search process can then be used to discover effective 
land-use policies for specific locations. Such a system was built on 
the Project Resilience platform \citep{itu:projectresilience23} and evaluated with the Land-Use 
Harmonization dataset LUH2 \citep{hurtt:luh20} and the bookkeeping model BLUE \citep{hansis:gbc15}. It generates Pareto
fronts that trade off carbon impact and amount of land-use change customized
to different locations, thus providing a proof-of-concept tool that is potentially useful for
land-use planning.
\end{abstract}

\section{Introduction}

One of the main factors contributing to climate change is CO$_2$ emissions due to land-use change \citep{friedlingstein:essd23}. Land-use CO$_2$ emissions depend crucially on how much
land area is allocated for different uses. Certain land-use changes such as afforestation can absorb carbon, and are considered a negative emission technology (NET) \cite{nap:net}. However not all land can be simply converted to forest; for example, land used for crops and pasture is essential for other purposes such as food supply. Land-use patterns must therefore be planned to minimize
carbon emissions while maintaining a satisfactory level of benefits received from the land, or ecosystem services. It is also important to note that land-use planning on its own is not enough to reach net-zero emissions. However, it can help offset emissions from sectors that are harder to decarbonize.

Recently, several different approaches were taken in order to optimize land-use planning to minimize carbon emissions. One such approach did so for the city of Nanjing in China using multi-objective linear programming \cite{cai:lumanage}. While effective, this optimization method cannot take advantage of nonlinearities. In another approach, multi-criteria decision-making (MCDM) was used to optimize urban development in the Rajshahi Metropolitan Area of Bangladesh \cite{rahman:gismcdm}. This approach uses manually ordered weights on outcomes to create solutions that trade off multiple objectives \cite{yager:owa}. Six pre-defined scenarios with fixed levels of trade-off are required in order to construct such weights. The land-use planning task was also represented as a causal machine learning task in the potential outcomes framework \cite{Rubin01032005} with crop type as the treatment, environmental factors as the features, and net primary productivity (NPP, i.e.\ the uptake flux in which carbon from the atmosphere is sequestrated by plants) as the outcome \cite{giannarakis2022assessingagriculturallandsuitability}. Causality between treatment and outcome was found to inform policy, but the approach does not create a comprehensive strategy. Thus, all three methods were effective in their approach but each had their own unique drawbacks.

In this paper, a new approach for land-use optimization was developed with these concerns in mind. It takes advantage of nonlinearities in the data and automatically generates a Pareto front of land-use planning policies that trade off multiple outcomes. This work was done as part of 
Project Resilience, a non-profit project hosted by the ITU agency of
the United Nations \citep{itu:projectresilience23}. The goal is to provide decision-makers with a tool
to know how their land-use choices affect CO$_2$ fluxes in the long term,
and make suggestions for optimizing these choices.

More specifically, the tool was designed to answer three questions: (1)
For a geographical grid cell identified by its latitude and
longitude, what changes to the land use can be made to reduce overall CO$_2$ emissions (or in some cases, maximize carbon absorption)?  (2) What will be the long-term CO$_2$ impact of changing 
land use in a particular way? (3) What are the optimal land-use
choices that can be made with minimal cost and maximal effect?

The approach is based on the Evolutionary Surrogate-assisted Prescription method \citep{francon:gecco20}. The idea is to first utilize historical data to learn a surrogate model on how land-use decisions in different contexts affect carbon emissions from land-use change, and then use this model to evaluate candidates in an evolutionary search process for good land-use change policies.

As a result, a Pareto front is generated of solutions that trade off
reduction in carbon emissions and the amount of change in land
use. Each point in the Pareto front represents an optimal policy for
that trade-off.  To make the results trustworthy, the tool allows the
decision-maker to explore modifications to these policies, and see the
expected effect. In the future, it should also be possible to evaluate
confidence of the predictions, evolve rule sets to make the policies
explainable, and utilize ensembling and further objectives and preferences
to make them more accurate.
Thus, the tool harnesses several techniques in machine learning to provide a proof-of-concept for decision-makers in optimizing land-use decisions.

The project page can be found at \url{https://project-resilience.github.io/platform/projects/landuse.html},
and an interactive demo at \url{https://landuse.evolution.ml}.

\section{Background}

The surrogate-assisted evolutionary optimization method used on the project is first outlined below, followed by the project background as the first application of Project Resilience.

\subsection{Evolutionary~Surrogate-assisted~Prescription}

\begin{figure}[t]
    \centering
    \includegraphics[width=0.35\linewidth]{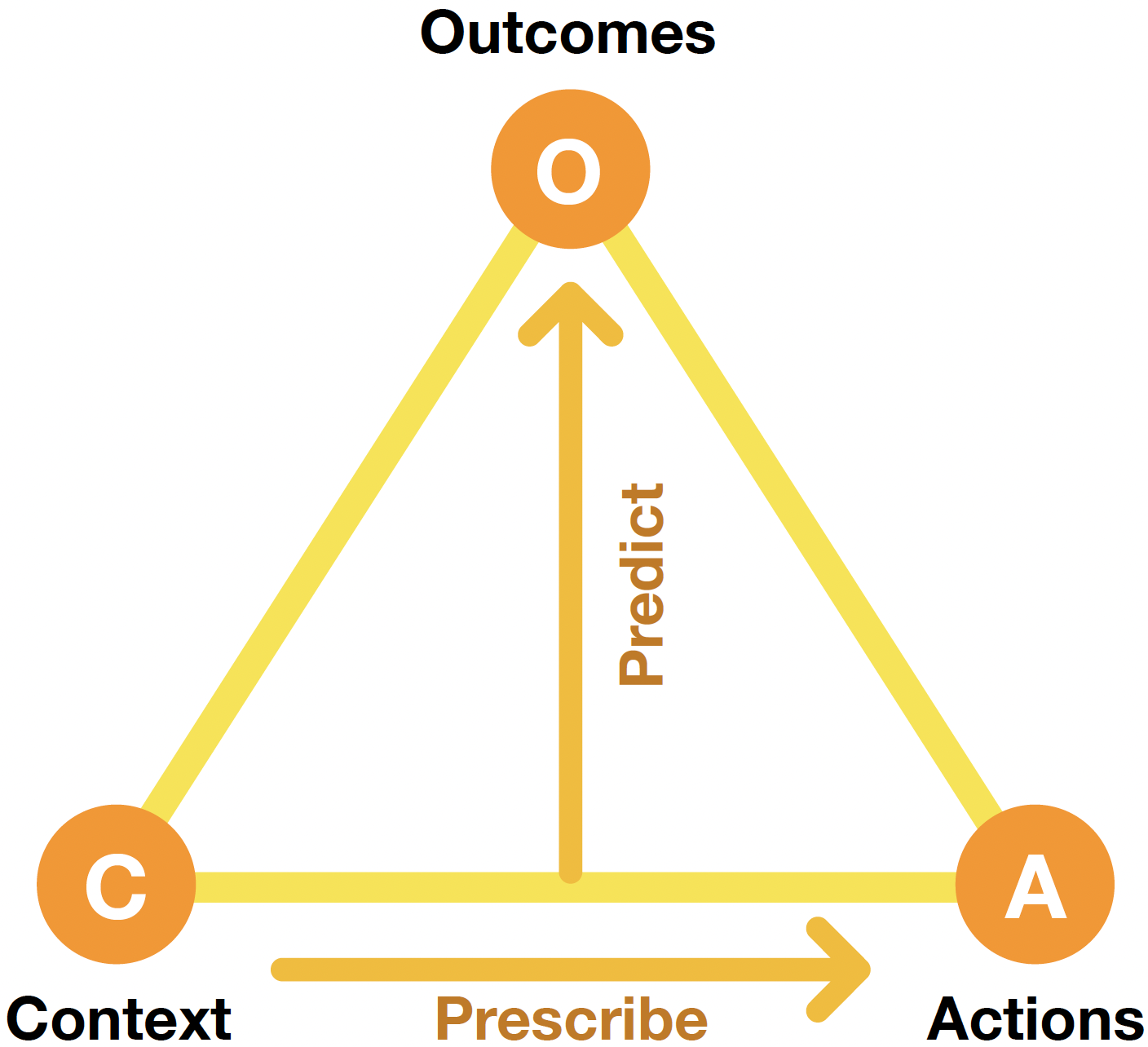}
    \caption{The Evolutionary Surrogate-assisted Prescription (ESP) method for decision optimization. A predictor is trained with historical data on how given actions in given contexts led to specific outcomes. It is then used as a surrogate in order to evolve prescriptors, i.e.\ neural networks that implement decision policies by prescribing actions for a given context, resulting in the best possible outcomes}
    \label{fg:esptriangle}
\end{figure}

Evolutionary Surrogate-assisted Prescription \citep[ESP;][]{francon:gecco20} is an approach for
optimizing decision-making in a variety of domains (Figure~\ref{fg:esptriangle}).
The main idea is that a decision policy can be represented as a neural network (or in some cases, as a set
of rules), and a good policy can be discovered through population-based
search, i.e.\ using evolutionary computation techniques, in order to prescribe actions in a given context. However, each
candidate must be evaluated, which is difficult to do in many
real-world applications. Therefore, a surrogate model of the world is
learned from historical data, predicting how good the resulting
outcomes are for each decision in each context.

More formally, given a set of possible contexts $\mathbb{C}$ and
possible actions $\mathbb{A}$, a decision policy $D$ returns a set of
actions $A$ to be performed in each context $C$:
\begin{equation}
D(C) = A\;,
\end{equation}
where $C \in \mathbb{C}$ and $A \in \mathbb{A}$. For each such $(C,A)$
pair there is a set of outcomes $O(C,A)$, and the Predictor $P_d$ is
defined as
\begin{equation}
P_d (C, A) = O,
\end{equation}
and the Prescriptor $P_s$ implements the decision policy as
\begin{equation}
P_s (C) = A\;,
\end{equation}
such that $\sum_{i,j} O_j(C_i,A_i)$ over all possible contexts $i$
and outcome dimensions $j$ is minimized (assuming decreased outcomes
are better). It thus approximates the optimal decision policy for the
problem. The predictor can be learned from historical data
on $CAO$ triples. In contrast, the optimal actions $A$ for each context
$C$ are not known, and must therefore be found through search.

More specifically in land-use optimization, context refers to the details of a given area such as current land use, a decision policy generates changes to the land use for the given area based on these details, and the outcomes are (1) the emissions from land-use change (ELUC) predicted by the surrogate as a result of this change, and (2) land-cost change, which is calculated as the percentage of the area changed. 

ESP was first evaluated in reinforcement learning tasks such as
function approximation, cart-pole control, and the Flappy Bird game,
and found to discover significantly better solutions, find them
faster, and with lower regret (deviation from the optimal reward) than standard approaches such as direct
evolution, Deep Q-Networks \cite[DQN;][]{mnih:nature15}, and
Proximal Policy Optimization \cite[PPO;][]{schulman:arxiv17}. 
It thus forms a promising foundation for optimizing land use as well.

\subsection{Project Resilience}

A major application of ESP to decision-making was developed for
optimizing strategies for non-pharmaceutical interventions (NPIs) in
the COVID-19 pandemic \citep{miikkulainen:ieeetec21}. Using case and NPI data from around the
world, collected in 2020-2022 by Oxford University's COVID-19
project \citep{hale:data20}, a neural network model was developed to predict the
course of the pandemic; this model was then used as a surrogate in
evolving another neural network to prescribe NPIs.  It discovered
different strategies at different phases of the pandemic, such as
focusing on schools and workplaces early on, alternating policies over
time, and focusing on public information and masking later. This
application demonstrated the power of ESP in discovering good
tradeoffs for complex real-world decision-making tasks.

Encouraged by this application, the XPRIZE Pandemic Response Challenge
\citep{xprize:pandemicresponse23} was developed in
2020-2021 for NPI optimization using the same predictor+prescriptor
approach. The competitors used several different machine learning or
other techniques, and drew significant attention to AI-supported
decision-making. The results were effective and in some cases used to 
inform decision-makers (e.g.\ on the 2021 surge in Valencia, Spain,
and on the fall 2021 school openings in Iceland). An important
conclusion was that it was possible to convene a group of experts---data scientists,
epidemiologists, and public health officials---to build a useful set of
tools to advise decision-makers on how to cope with and plan around a health
disruption to society.

This experience with the XPRIZE Pandemic Response Challenge then led to
Project Resilience \citep{itu:projectresilience23}. The XPRIZE competition serves as a
blueprint for a collaborative effort to tackle global problems. The
goal of Project Resilience is to build a public AI utility where a
global community of innovators and thought leaders can enhance and
utilize a collection of data and AI approaches to help with better
preparedness, intervention, and response to environmental, health,
information, social equity, and similarly scoped problems. The project
is hosted by the ITU agency of the United Nations, under its Global
Initiative for AI and Data Commons, and contributes to its general
efforts toward meeting the Sustainable Development Goals (SDGs).

Staffed by volunteers from around the world, Project Resilience was
started in 2022 with the design of a minimum viable product (MVP) platform. The goal is to
\begin{quote}
\begin{itemize}
\item Develop an architecture to pull input and output data hosted
  by third parties;
\item Build an application programming interface (API) for third parties to submit models.
\item Develop code to compare both predictors and prescriptors in
  third-party models and produce a set of performance metrics;
\item Build a portal to visualize the assessment of predictors and
  prescriptors to include generations of key performance indicators
  (KPIs) and comparison across models;
\item Develop an ensemble model for predictors and prescriptors; and
\item Provide a user interface to decision-makers to help them in 
  their decision-making process.
\end{itemize}
\end{quote}
Land-use optimization is the first application of this MVP,
as described below.

\vspace*{-1ex}
\section{Land-Use Optimization Task}

The data sources are described first, followed by how they are used as
the context, action, and outcome variables. The elements of the
approach include the predictor and prescriptor models, a method for
modeling uncertainty, and ensembling the results.

\vspace*{-1ex}
\subsection{Data}

The data for carbon emissions (Emissions resulting from Land-Use Change, ELUC) originate from the Bookkeeping of Land-Use Emissions model \citep[BLUE;][]{hansis:gbc15}, which attributes carbon fluxes to land-use activities. A bookkeeping model is a semi-empirical statistical model that calculates CO$_2$ fluxes based on changes in the carbon content of biomass and soil caused by land-use changes. Bookkeeping models employ different carbon densities for different vegetation types (plant functional types in BLUE) and use response functions to determine the temporal dynamics of the fluxes. Advantages of bookkeeping models include their ability to separate anthropogenic from natural CO$_2$ fluxes and to attribute fluxes to specific land-use change events \cite{dorgeist:nature}. Bookkeeping models thus differ from more general climate models and land surface models that are based on parameterizations of physical, chemical, and biological processes. BLUE is one of four bookkeeping models that are used to provide ELUC estimates for the annually published Global Carbon Budget \cite[GCB;][]{friedlingstein:essd22,friedlingstein:essd23,friedlingstein:essd24}. BLUE is particularly useful because it provides spatially explicit ELUC estimates. In this paper,  BLUE is used to estimate committed emissions due to land-use changes, which means that all the emissions or removals caused by
a land-use change event are attributed to the year of the event. The emission (or removal) estimates thus relate to the long-term impact of a land-use change event, i.e.\ up to multiple decades \citep{hansis:gbc15,houghton:ecomono83}.
While in principle BLUE can be used as the surrogate model for ESP,
in practice its calculations are too expensive to carry out on 
demand during the search for good policies. Therefore, 
a number of BLUE simulations were performed covering a comprehensive set of
land-use changes for 1850-2022, resulting in a dataset that could
be used to train an efficient surrogate model.

The Land-Use Change (LUC) data is provided by the Land-Use Harmonization project \citep[LUH2;][]{hurtt:luh20}. LUH2 provides data on fractional land-use patterns, underlying land-use transitions, and key agricultural management information, annually for the time period 850-2100 at 0.25$\times$0.25 degree resolution. LUH2 provides separate layers for states, transitions, and management. Version LUH2-GCB2022 from the Global Carbon Budget 2022 \citep{friedlingstein:essd22} was the basis for this paper. It is based on the HYDE3.3 historical database \cite{kleingoldewijk:essd17,kleingoldewijk:jlus17} and uses the 2021 wood harvest data provided by the FAO agency of the United Nations. It consists of the following land-use types:

\begin{quote}
\begin{description}
\item
\hspace*{-1.1ex}Primary: Vegetation that has never been impacted by human activities (e.g. agriculture or wood harvesting) since the beginning of the simulation
\vspace*{-1.35ex}
\begin{itemize}
\item
primf: Primary forest
\item
primn: Primary nonforest vegetation
\end{itemize}

\vspace*{0.5ex}
\item
\hspace*{-1.1ex}Secondary: Vegetation that is recovering from previous human disturbance (either wood harvesting or agricultural abandonment)
\vspace*{-1.35ex}
\begin{itemize}
\item
secdf: Secondary forest
\item
secdn: Secondary nonforest vegetation
\end{itemize}

\item
\hspace*{-1.1ex}Urban
\vspace*{-1.35ex}
\begin{itemize}
\item
urban: Urban areas
\end{itemize}

\item
\hspace*{-1.1ex}Crop
\vspace*{-1.35ex}
\begin{itemize}
\item
crop: All crop types pooled into one category
\end{itemize}

\item
\hspace*{-1.1ex}Pasture
\vspace*{-1.35ex}
\begin{itemize}
\item
pastr: Managed pasture land
\item
range: Grazed natural land (i.e., grazing on grassland, savannah, etc.).
\end{itemize}
\end{description}
\end{quote}
\vspace*{1ex}

BLUE calculates yearly ELUC estimates based on the land-use changes provided by LUH2. Urban areas are pooled into a single type; their carbon contents are too low and the total urban area on the planet is too small for differences among them to have a significant effect on the CO$_2$ flux. All crop types are also pooled into a single crop category. BLUE considers that the degradation of primary to secondary land often lowers the carbon content in vegetation and soils, and it considers pastures and the degradation of natural grasslands and savannahs by grazing. A complete overview of the model can be found in \citep{hansis:gbc15}.

It is important to note two caveats in this data. First, the 0.25-degree spatial resolution of LUH2 is rather coarse for land-use recommendations. Second, different crop types do have noticeable differences when making decisions at a local level. To address these concerns, in the future more fine-grained land-use estimates can be used and crop type can be added as a feature variable, and such extensions would allow for more accurate prescriptions at the local level.  In the meantime, the current versions of LUH2 and BLUE are sufficient to build a proof-of-concept system that demonstrates the power of land-use planning through neuroevolution.

Together the ELUC and LUC datasets form the basis for constructing the
context, action, and outcome variables, as will be described next.

\subsection{Decision-making Problem}
\label{sc:problem}

The modeling approach aims to understand the domain in two ways: (1) In a
particular situation, what are the outcomes of the decision maker's
actions?  (2) What are the actions that result in the best outcomes,
i.e.\ the lowest ELUC with the smallest change in land use? The data is thus organized into context, action, and outcome variables.


{\bf Context} describes the problem the decision maker is facing, i.e.\
a particular grid cell, a point in time when the decision has to be 
made, and the current land use at that time. More specifically it consists of:

\begin{quote}
\begin{itemize}
\item Latitude and Longitude, representing the cell on the grid. Climate information loosely corresponding to plant functional types is implicitly represented in the location of the cell, as shown in Appendix \ref{appendixA}.
\item Area, representing the surface of the cell. Cells close to the
  equator have a bigger area than cells close to the poles.
\item Year, useful to capture historical decisions: The same cell
  may have been through many land-use changes over the years.
\item Land use, representing the percentage of land in the cell used by each type above. For simplicity, all crop types were pooled into a single type. The percentages do not necessarily reach 100\% in each grid cell; a 'nonland' type (e.g.\ typically sea, lake, etc.) is assumed to fill the remainder.
\end{itemize}
\end{quote}


{\bf Actions} represent the choices the decision-maker faces. How can they
change the land? In the study of this paper, these decisions are
limited in two ways:

First, decision-makers cannot affect primary land. The idea is that it
is always better to preserve primary vegetation, such as pristine rainforests; destroying it is not
an option given to the system. Technically, it is not possible to
re-plant primary vegetation. Once destroyed, it is destroyed
forever. If re-planted, it would become secondary vegetation, for which BLUE assumes lower carbon content than for primary vegetation \cite{hansis:gbc15}.

Second, decision-makers cannot affect urban areas. The
needs of urban areas are dictated by other imperatives, and optimized
by other decision makers. Therefore, the ESP system cannot recommend that
a city should be destroyed, or expanded. Further constraints on crop usage for food production are discussed in Section \ref{sc:crop}


{\bf Outcomes} consist of two conflicting variables. The primary
variable is ELUC, i.e.\ emissions from land-use change. It consists of
all CO$_2$ emissions attributed to the land-use change, in metric tons of carbon per hectare (tC/ha), obtained
from the BLUE simulation. A positive ELUC means carbon is emitted to the atmosphere and a negative number means carbon is taken up by vegetation. For simplicity, all future CO$_2$ emissions or removals are attributed to the initial year of the land-use change; temporal dynamics in CO$_2$ fluxes are not taken into account. Similarly, the full land-use change is assumed to take place; incremental changes are not taken into account. The secondary variable is the cost of the land-use 
change, represented by the percentage of land that was changed. This
variable is calculated directly from the actions.

There is a tradeoff between these two objectives: It is easy to reduce
emissions by changing most of the land, but that would come at a huge
cost. Therefore, decision-makers have to minimize ELUC while
minimizing land-use change at the same time. Consequently, the result is not
a single recommendation, but a Pareto front where each point
represents the best implementation of each tradeoff given a balance between the two
outcomes.

\section{Models}

The system consists of the predictor, trained with supervised learning
on the historical data, and the prescriptor, discovered through a evolutionary search in which each candidate is evaluated with the predictor.


{\bf Prediction:} Given the context and actions that were performed, the predictive
model estimates the outcomes. In this case, since the cost outcome (i.e.\ the percentage of changed land) can
be calculated directly, only the ELUC is predicted by the model. That
is, given the land use of a specific location, and the land-use changes that
were made during a specific year, the model predicts the long-term CO$_2$
emissions directly caused by these land-use changes. The model does not predict non-ELUC emissions such as forest fires that may lead to a change into secondary forest.

Any predictive model can be used in this task, including linear regression, random forest, or a neural network; they are each evaluated below. As usual in machine learning, each model is fit to the existing historical data and evaluated with left-out data.


{\bf Prescription:} Given context, the prescriptive model suggests actions that optimize
the outcomes. The model has to do this for all possible contexts, and
therefore it represents an entire strategy for optimal land use.
The strategy can be implemented in various ways, like sets of rules or neural networks. The approach in this paper
is based on neural networks.

The optimal actions are not known, but the performance of each
candidate strategy can be measured (using the predictive model), therefore the
prescriptive model needs to be learned using search techniques.
Standard reinforcement learning methods such as PPO and DQN are
possible; the experiments in this paper use evolutionary
optimization, i.e.\ conventional neuroevolution \citep{stanley:naturemi19}.  As in prior applications of ESP \citep{francon:gecco20, miikkulainen:ieeetec21}, the network has a fixed architecture of two fully connected layers; its weights are
concatenated into a vector and evolved through crossover and mutation.

\section{Experiments}

This section describes the main results of the project so far, including different ways to create good predictors and the evolution of effective prescriptors based on them. The behavior of the discovered prescriptors is analyzed quantitatively and through various ablation and extension studies. As a concrete implementation of the mechanisms, an interactive demo of the entire land-use recommendation system is described.

\subsection{Prediction}

In preliminary experiments, prediction performance was found to differ between major geographical regions. To make these differences explicit, separate models were trained on different subsets of countries: Western Europe (EU), South America (SA), and the United States (US), making up 1.12, 9.99, and 7.26 percent of the global dataset, respectively. The EU subset contains the UK, France, Germany, the Netherlands, Belgium, Switzerland, and Ireland. The SA subset includes Brazil, Bolivia, Paraguay, Peru, Ecuador, Colombia, Venezuela, Guyana, Suriname, Uruguay, Argentina, and Chile. These countries were chosen as a representative subset to make the experiments computationally less expensive.

Upon examining the global dataset, a linear relationship was found between $|$ELUC$|$ and land-use change with an R-squared value of 0.92. Therefore, a linear regression (LinReg) was fit to the data as the first predictive model. In preliminary experiments, the LinReg model was found to perform the best when current land use, latitude, longitude, cell area, and time were removed from the inputs. Thus, the difference in land use was used as the only input feature in the experiments. Data from the years 1851--2011 (inclusive), totaling 38,967,635 samples globally, was used to train the LinReg models; the 2,420,350 remaining samples from 2012--2021 (inclusive) were used for testing.

Second, a random forest model (RF) was trained on the same data but now includes the full set of input features, in order to take advantage of RF's ability to decide which features are most useful. The forest consisted of 100 decision trees with unrestricted maximum depth; at each split, a random subset of $n$ features was considered (where $n$ is the square root of the total number of features). To make the training time of the Global RF model feasible, it was trained with data from only 1982--2011. Preliminary experiments showed that constraining the training set in this manner did not decrease performance significantly.  Both the linear regression and random forest models were trained using the scikit-learn library \cite{scikit-learn}.

Third, a neural network (NeuralNet) was trained for the same task with the full set of features and data from 1851--2011 using the PyTorch library \cite{paszke2019pytorch}. The network was fully connected, composed of an input layer, a single hidden layer of size 4096, and an output layer. The output layer took as input the concatenation of the input layer and the hidden layer in order to take advantage of the observed linear relationships more easily \citep{cheng2016wide}. It was trained for three epochs with batch size 2048 using the AdamW optimization algorithm \cite{loshchilov:iclr19}. After each epoch, the learning rate was scaled by a factor of 0.1.

\begin{table}[!p]
\centering
\small
\setlength{\tabcolsep}{1.25ex}
\begin{tabular}{c | r | l l l l}
    \textbf{Model} & Time (s) & EU & SA & US & Global \\
    \hline
LinReg (EU) & 0.047 & 0.033 & 0.172 & 0.169 & 0.206\\
LinReg (SA) & 0.457 & 0.137 & 0.153 & 0.061 & 0.110\\
LinReg (US) & 0.331 & 0.139 & 0.146 & 0.035 & 0.073\\
LinReg (Global) & 4.644 & 0.139 & 0.150 & 0.035 & 0.074\\
\hline
RF (EU) & 17.697 & $0.064$ & $0.211^{\dag}$ & $0.161^{\dag}$ & $0.218^{\dag}$\\
RF (SA) & 209.688 & $0.133^{*\dag}$ & $\textbf{0.071}^{*\dag}$ & $0.074^{\dag}$ & $0.126^{\dag}$\\
RF (US) & 111.701 & $0.163$ & $0.185^{\dag}$ & $0.032^{*}$ & $0.094^{\dag}$\\
RF (Global) & 417.647 & $0.041^{*\dag}$ & $0.076^{*\dag}$ & $0.028^{*}$ & $\textbf{0.045}^{*\dag}$\\
\hline
NeuralNet (EU) & 10.711 & $\textbf{0.025}^{*\dag}$ & $0.277$ & $0.286$ & $0.334$\\
NeuralNet (SA) & 103.696 & $0.248$ & $0.100^{*}$ & $0.562$ & $0.399$\\
NeuralNet (US) & 73.141 & $0.136^{\dag}$ & $0.225$ & $\textbf{0.024}^{*\dag}$ & $0.150$\\
NeuralNet (Global) & 1649.193 & $0.046^{*}$ & $0.110^{*}$ & $0.025^{*\dag}$ & $0.050^{*}$\\
\hline
\end{tabular}\\
\vspace*{0.25ex}
{\footnotesize {\bf Bold}: the best model in each region; *: 99\% confidence the model outperforms LinReg;\\[-1ex]
\dag: 99\% confidence that RF outperforms NeuralNet, or vice-versa}
\vspace*{1ex}
\caption{Average training time and mean absolute error (MAE) in tC/ha of the trained models of each type trained on each region and evaluated on that region as well as all other regions. Statistical significance was estimated in each case with the unpaired $t$-test. The NeuralNet models are more accurate RF on EU and US, RF more accurate than NeuralNets on SA and globally; both are more accurate than LinReg in all regions (except RF on EU). Overall, the specialized models are more accurate than the Global models. However, RF models do not extrapolate well (as seen in Figure~\ref{fg:all-model-heatmaps}), which is necessary for evaluating prescriptor candidates. In contrast, NeuralNets do, and they are also otherwise reasonably accurate. Therefore, the Global NeuralNet model was used to evolve the prescriptors}
\label{tb:predictors}
\end{table}

\vspace{1ex}

\begin{figure*}[!p]
    \centering
    \includegraphics[width=\linewidth]{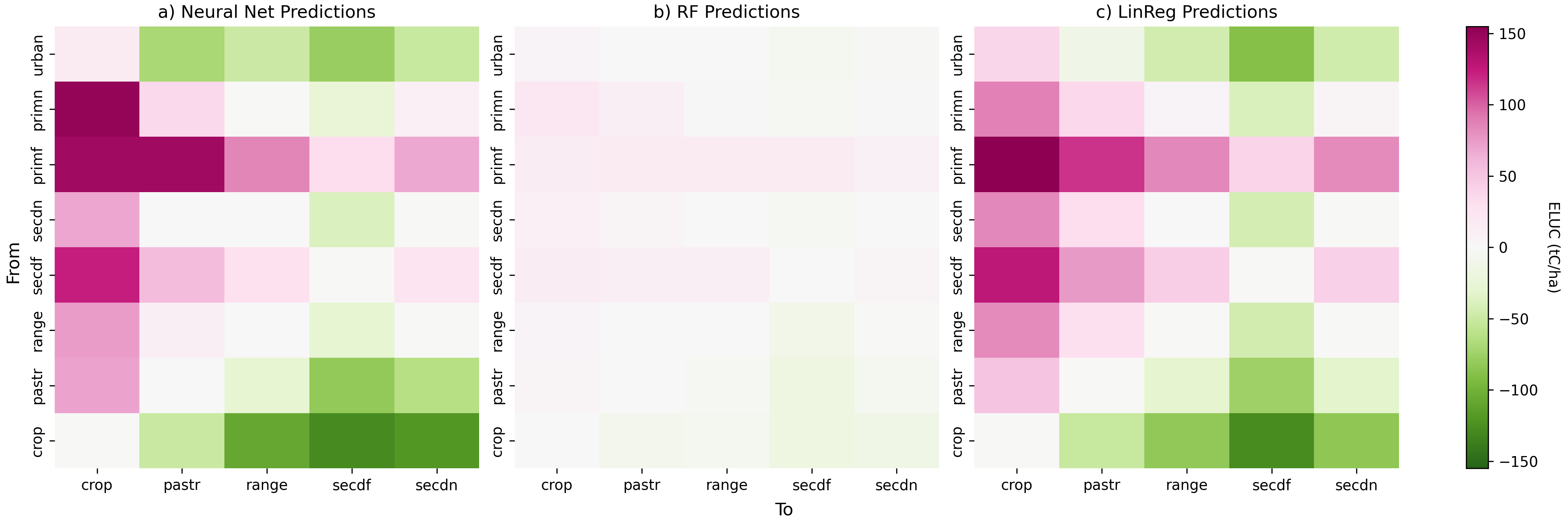}\\[-1ex]
    \caption{Visualizing the differences in model behavior. Predictions for ELUC (tC/ha) are created for the Global models using synthetic data created by changing 100\% of land-use type A (row) to 100\% of type B (column) in a 1\% sample across the range of latitude, longitude, cell area, and year occurring in the test data. The results are averaged for each conversion from A to B. The models generally agree on the sign of ELUC, which in turn aligns with the sign generated by the BLUE model, suggesting that the results are reliable. The RF model is not able to extrapolate to large values, resulting in low predictions; LinReg and NeuralNet are similar but differ numerically, presumably due to the differences between linear and nonlinear predictions}
    \label{fg:all-model-heatmaps}
\end{figure*}

LinReg, RF, and NeuralNet models were trained separately on each region and globally. While LinReg is deterministic and needed to be trained only once, RF and NeuralNet are stochastic in their initialization and training and were therefore trained 30 times to evaluate statistical significance. All models were tested on all regions with data from the most recent years, i.e.\ 2012--2021, and evaluated on mean absolute error (MAE) of predicted ELUC.  As shown in Table~\ref{tb:predictors}, the LinReg models were fast to train, but they performed the worst (i.e., they have the largest mean absolute error in almost all cases), suggesting that the linear relationship is not enough to account for the data fully. RF performed significantly better than LinReg in all regions but EU, significantly worse than NeuralNet in EU and US, and outperformed NeuralNets in SA and globally. However, RF models do not generalize/extrapolate well to large changes in land use that occur in the dataset infrequently (or not at all for some land types; Figure~\ref{fg:all-model-heatmaps}), which turns out to be important when evolving prescriptors. Many prescriptor candidates suggest large land-use changes, and the predictor model needs to be able to evaluate them. In contrast, NeuralNets capture nonlinearities, which explains why they outperform LinReg in all regions. Most importantly, they extrapolate well, which suggests that they are a better choice for the surrogate model in prescriptor evolution. Additionally, LinReg and NeuralNet models have many fewer parameters than RF (taking 4KB and 363KB on disk respectively vs.\ 14GB for RF).

In order to visualize the behavior of the predictor models, they were queried with a synthetic dataset created with every possible conversion from 100\% land type A to 100\% land type B (the types primf, primn, and urban were excluded from B as was explained in Section~\ref{sc:problem}). Because the RF and NeuralNet take latitude, longitude, cell area, and year as input, 24,204 (1\%) latitude/longitude/cell-area/year combinations from the test dataset were sampled randomly and evaluated with each conversion. The results were averaged across the samples to get the overall result for the conversion. Figure~\ref{fg:all-model-heatmaps} shows these results for the Global models. The first finding is that all models generally rank conversions to crops $<$ pasture $<$ forest in terms of reducing ELUC, and generally agree on the sign of ELUC. This ranking also generally aligns with those generated by the BLUE model. This consistency suggests that the results are reliable. Second, while LinReg and NeuralNet make similar predictions overall, their numerical values differ in many places, presumably due to the nonlinearities. Third, despite RF's generally high accuracy in known scenarios, it is not able to generalize/extrapolate to large changes and instead predicts low-magnitude ELUC values for all imposed land-use transitions.

Thus, since NeuralNet achieves similar or better performance than the RF, extrapolates well to extremes, and can learn nonlinear relationships in the data, the Global NeuralNet was used as the surrogate model to evolve the prescriptors.

\vspace*{3ex}
\subsection{Prescriptions}
\label{prescriptions-section}

The prescriptors were fully connected neural networks with two layers of weights, implemented with PyTorch. The topology was fixed and only the weights were evolved. The input consisted of 12 float values, i.e.\ the eight land-use percentages and the latitude, longitude, and area of the geographical cell, as well as the year. The hidden layer contained 16 units with tanh activation.

During evolution, prescriptor candidates were evaluated on a fixed random subset of 1\% of the Global dataset from years 1851--2011 with the Global NeuralNet predictor. Prescriptor output was a single vector of size five (the eight land-use percentages minus primf, primn, and urban, which were considered fixed), produced by the ReLU activation function. These five outputs were normalized to sum to the previous amount of land taken by these five types. The result is the prescriptor's suggestion for the percentage of land use of each type. The five outputs were combined with the fixed land-use values for primf, primn, and urban to compute the difference from the initial land use. This difference was concatenated with the initial land use as well as latitude, longitude, cell area, and year, and passed to the Predictor model to get the predicted ELUC metric. The cost outcome, i.e.\ the percentage of land-use change, was computed based on the difference between the five outputs and the corresponding inputs. All positive differences were summed for a given cell and divided by the total land-use percentage of the cell (excluding nonland). These metrics were aggregated across the evaluation data set to obtain the performance outcome of each prescriptor candidate. Thus, a candidate was evaluated based on its average performance in creating a strategy for each land area in the test set.

\begin{figure*}[t]
    \centering
    \includegraphics[width=\linewidth]{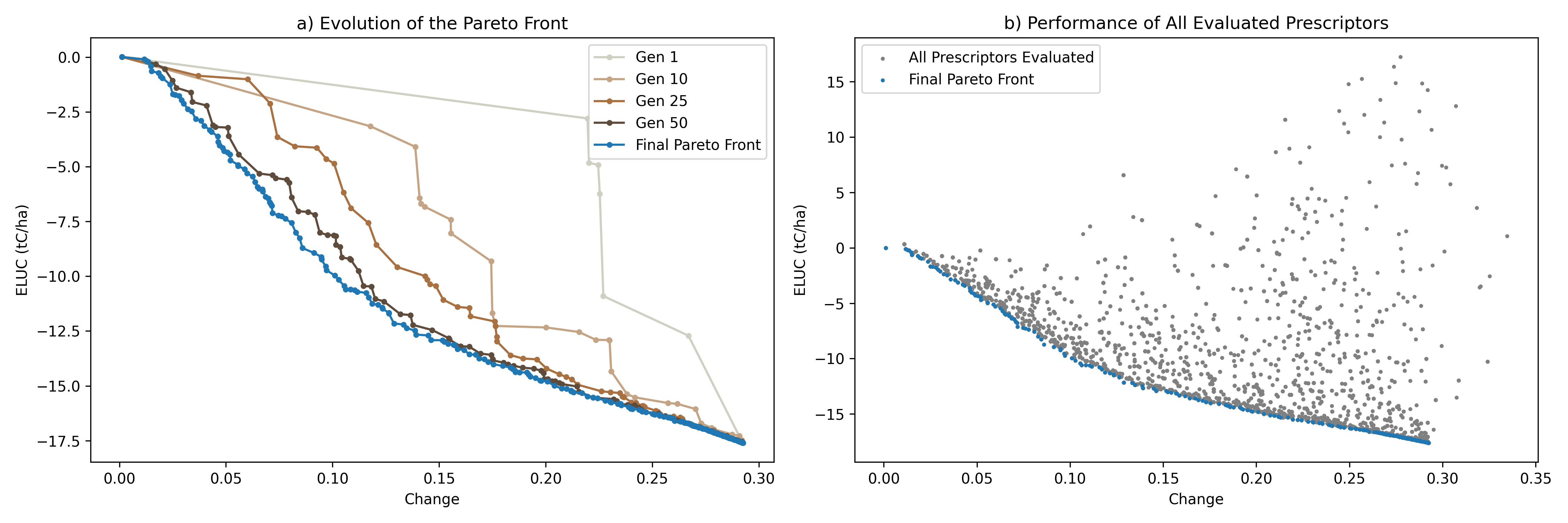}\\[-1ex]
    \caption{Evolution of prescriptors with the Global NeuralNet predictor. ($a$) The Pareto front moves towards the lower left corner over evolution, finding better implementations for the different tradeoffs of the ELUC and land-use change amount objectives. ($b$) Each prescriptor evaluated during evolution is shown as a dot, demonstrating a wide variety of solutions and tradeoffs. The final Pareto front (collected from all generations) is shown as blue dots. It constitutes a set of solutions from which the decision-maker can choose a preferred one}
    \label{fg:nn-global-allprescriptors}
\end{figure*}

The aggregated ELUC and the land-use change outcomes were minimized with an evolutionary search on the neural network weights. Conventional neuroevolution with crossover and mutation operators on a vector of network weights was combined with NSGA-II \citep{deb:ppsn00} to make the search multiobjective. In order to seed evolution with good coverage of the search space, two prescriptor models were trained with backpropagation and injected into the initial population: One to prescribe no land-use changes at all and the other to prescribe as much secondary forest (secdf) as possible (which was observed to reduce ELUC the most). This process takes advantage of the Realizing Human Expertise through AI (RHEA) framework \cite{meyerson:neurips24} for distilling human intuition into evolutionary search. Details and the impact of this seeding mechanism are discussed in Section \ref{sc:rhea}. The rest of the population was created randomly through an orthogonal initialization of weights in each layer with a mean of zero and a standard deviation of one \citep{saxe2013orthogonalinit}. Evolution was run for 100 generations with the following parameters:
\begin{quote}
\begin{itemize}
\item mutation\_type: gaussian\_noise\_percentage
\item mutation\_factor: 0.1
\item mutation\_probability: 0.2
\item population\_size: 100
\item parent\_selection: tournament
\item initialization\_distribution: orthogonal.
\end{itemize}
\end{quote}

Figure~\ref{fg:nn-global-allprescriptors} demonstrates the progress of evolution towards increasingly better prescriptors, i.e.\ those that represent better implementations of each tradeoff of the ELUC and land-use change objectives. These solutions are referred to as "Evolved Prescriptors" in the rest of the paper. They represent a wide variety of tradeoffs and a clear set of dominant solutions that constitute the final Pareto front. That set is returned to the decision-maker, who can then select the most preferred one to implement.

\begin{figure}[t!]
    \centering
    \includegraphics[width=0.49\linewidth]{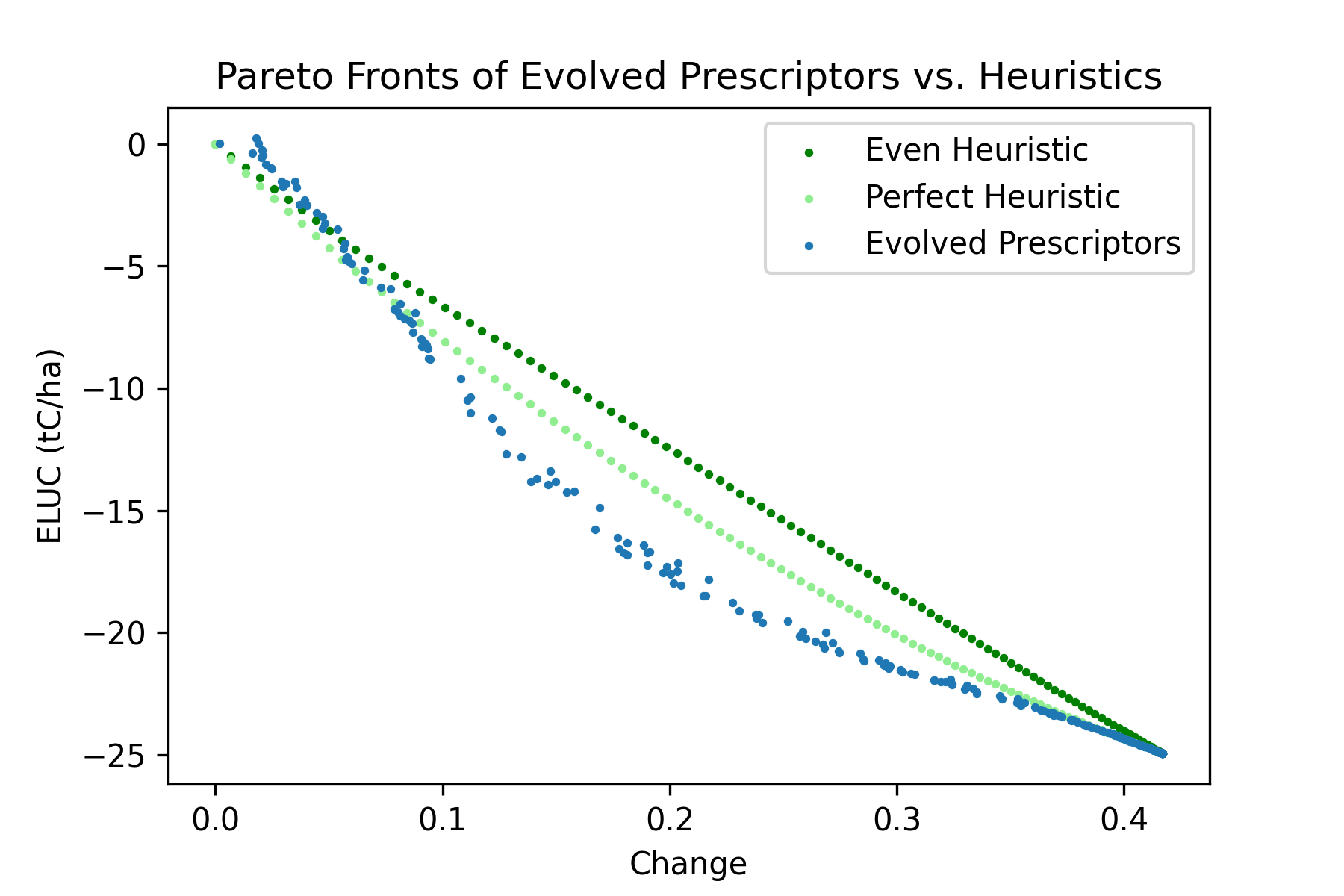}\\[-1ex]
    \caption{The Pareto fronts of Evolved Prescriptors vs.\ heuristic baselines, with ELUC and land-use change evaluated on the Global test set. The Evolved Prescriptors achieved better solutions than the baselines in the middle-change region where the land-use changes matter the most, demonstrating that they can take advantage of nonlinear relationships in land use to discover useful, non-obvious solutions}
    \label{fg:heuristics}
\end{figure}

To demonstrate the value of evolutionary search, the Evolved Prescriptors were compared against two heuristic baselines (Figure~\ref{fg:heuristics}). Each of them converts as much land as possible to secdf, which has the lowest ELUC, up to a set threshold. The first is a simple baseline called "Even Heuristic", where an equal proportion of land up to the threshold is taken from all land-use types and converted to secdf. The second is an optimal linear baseline called "Perfect Heuristic", where as much land as possible is first taken from the land-use type that has the highest weight in the Global LinReg model, then from the land-use type that has the second highest weight, and so on until either the given land-use change threshold is reached or there is no land left to convert to secdf.  These heuristic solutions were created with evenly distributed land-use change thresholds between 0\% and 100\% in order to generate heuristic Pareto fronts.  All three methods were then evaluated on a random 1\% sample of the Global test set from the years 2012--2021.

As Figure~\ref{fg:heuristics} shows, the Evolved Prescriptors and the heuristics all converged in the high-change region of the Pareto front, where 100\% forest prescriptions dominate. The heuristics performed slightly better in the low-change region, where the predictors may not be accurate enough to distinguish between small changes. However, in the most useful middle-change region, the Evolved Prescriptors outperformed both heuristics, demonstrating that they are likely exploiting the nonlinear relationships well. 

The performance of each prescription method can be quantified in terms of hypervolume, which measures how much of the solution space it dominates. It is calculated as the area above and to the right of the Pareto front up to maximum possible change and zero ELUC, divided by the area of the entire search space (which starts from zero change and lowest possible ELUC, which in turn is defined as the ELUC of the Perfect Heuristic at the maximum change). The Evolved Prescriptors outperform the Perfect Heuristic, which outperforms the Even Heuristic, with hypervolumes of 0.616, 0.566, and 0.512, respectively.

\vspace*{3ex}
\subsection{Prescriptor Behaviors}


\begin{figure*}[t!]
    \centering
    \includegraphics[width=\linewidth]{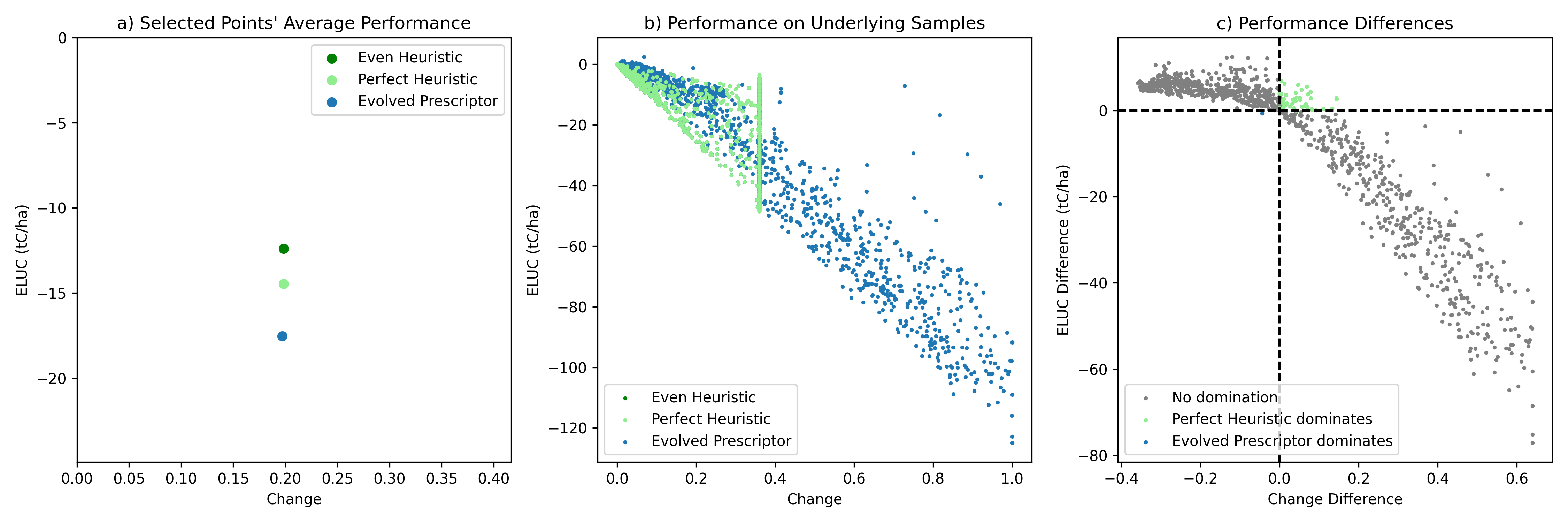}\\[-1ex]
    \caption{Comparing a selected Evolved Prescriptor with the Perfect Heuristic. ($a$) The average performance of the Evolved Prescriptor dominates that of the heuristic. ($b$) The averages are expanded into actual samples in the test set (subsampled for readability). The samples for the Even Heuristic are largely hidden under the samples for the Perfect Heuristic. The Evolved Prescriptor suggests many more large changes than the heuristic. ($c$) The differences in change percentage and ELUC between the Evolved Prescriptor and the Perfect Heuristic for each test sample, with color indicating which solution dominates. Surprisingly, this particular Evolved Prescriptor dominates the Perfect Heuristic only on a single individual sample. Thus, evolution discovered the insight that it is possible to do well globally by utilizing a few cases where large change is possible}
    \label{fg:presc-comparison}
\end{figure*}

It is interesting to analyze how the Evolved Prescriptors were able to improve upon the linear heuristics, which were constructed to be quite good to begin with. One representative from each method was chosen from the middle-change region and compared (Figure \ref{fg:presc-comparison}$a$). The heuristic methods both prescribe 19.8\% change and -12 and -14 ELUC on average, and the closest Evolved Prescriptor that dominates them averages 19.7\% change and -18 ELUC.  In Figure \ref{fg:presc-comparison}$b$ these averages were expanded to show the results for actual samples in the test set (only 10\% of those samples are plotted for readability). The samples for the Even Heuristic are hidden under those of the Perfect Heuristic: When these heuristics cannot reach their allocated thresholds, they change everything to secondary forest, resulting in overlapping points; their points similarly overlap at the vertical line indicating the change threshold. Note that the Evolved Prescriptor allocates more change to several samples than the Heuristic: This is because the heuristic's change threshold is set to 0.35 (in order to get to $\sim$20\% change on average) whereas the Evolved Prescriptor has no such restriction.

The differences in change percentage and ELUC between the Perfect Heuristic and a typical Evolved Prescriptor are plotted for each sample in the test set in Figure \ref{fg:presc-comparison}$c$. The heuristic dominates in the samples in the upper right quadrant (where both differences are above 0, i.e.\ the prescriptor's change and ELUC are both larger than those of the heuristic) and the prescriptor in the bottom left quadrant. Surprisingly, the Evolved Prescriptor dominates the Perfect Heuristic in only 1 of the 24,204 test samples. How can this be?

The reason is that evolution discovered a way to take advantage of large changes. As shown in Figure~\ref{fg:presc-comparison}$b$, the trained prescriptor is not constrained by a limit on every change it makes: As long as its changes remain below the limit on average, it can make a few large changes. 
As a result, evolution discovered a strategy where it makes only a few large changes, and makes them in cells where they have the largest impact. Thus, even though the Evolved Prescriptor almost never beats the Perfect Heuristic in any individual cell, when averaged across all cells, it dominates the heuristic. In other words, pick your battles right, and you'll win overall.

\begin{figure}[t]
    \centering
    \includegraphics[width=\linewidth]{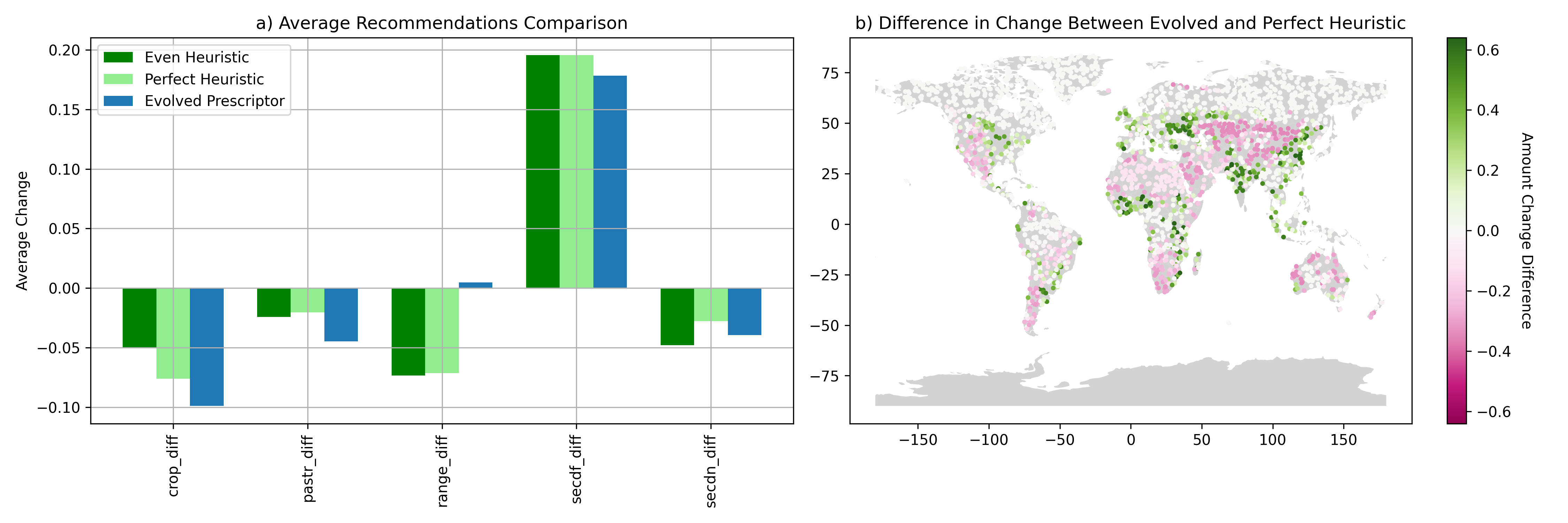}\\[-1ex]
    \caption{Characterizing the recommendations. ($a$) The strategies used by the Evolved Prescriptors and the heuristics are illustrated by plotting the average amount of change in each land-use type. Primarily, they all suggest converting cropland to secondary forest. The Evolved Prescriptor removes more crop on average, and generally takes advantage of more change. ($b$) The learned ability of the Evolved Prescriptor to allocate more or less change to certain regions. Primary forest is left untouched, dry regions are changed less, and tropical, temperate, and continental regions are changed more
    }
    \label{fg:presc-explanation}
\end{figure}

In order to see what the recommendations look like, the average policies of the three approaches are illustrated in Figure \ref{fg:presc-explanation}$a$. The greatest amount of negative change made is to crop and positive is to secondary forest, i.e.\ all approaches recommend converting cropland to forest, which makes sense. The greatest departure from the heuristics is that the Evolved Prescriptors recommend more crop change and almost no rangeland change.

Where the Evolved Prescriptors picked their battles compared to the Perfect Heuristic is shown in Figure \ref{fg:presc-explanation}$b$. Certain areas such as the Amazon rainforest were left untouched by both, 
because the system does not allow primary forests (that capture much carbon) to be changed. The Evolved Prescriptors avoided change in dry regions, such as the US Southwest, Sahara, Middle East, Central Asia, and Northwestern Australia, where presumably only limited change is 
effective. The prescriptors recommended more change in tropical, temperate, and continental regions like the US Midwest, Gulf of Guinea, South and Southeast Asia, Continental Europe, and the Pampas region of South America. These locations align with the change profile in Figure \ref{fg:presc-explanation}$a$: The areas where more crops can grow are changed more, whereas plains and desert regions corresponding to rangeland are barely touched.



\subsection{Constraining Crop Loss}\label{sc:crop}

While minimizing ELUC and change allows a decision-maker to find solutions along an important trade-off, it does not limit how much cropland the prescriptor is allowed to change, which could lead to food shortages. A step towards a more practical application is to optimize under such a constraint. To this end, the same evolutionary setup was run with a third objective of minimizing absolute change in crops. This approach assumes that the current level of crops is sufficient, however, the constraint could easily be set to an arbitrary value. Consequently, a seed model, trained through backpropagation, was included in the initial population to change as much land as possible to forest without removing any cropland. Such a seed makes it more likely for evolution to discover solutions that do not prescribe as much crop change. Evolution parameters were the same as in Section \ref{prescriptions-section}. These newly evolved prescriptor models were compared to the previous heuristics as well as to a new "No Crop" Heuristic, i.e. the Perfect Heuristic without modifying any cropland.

\begin{figure}[!t]
    \centering    
    \includegraphics[width=\linewidth]{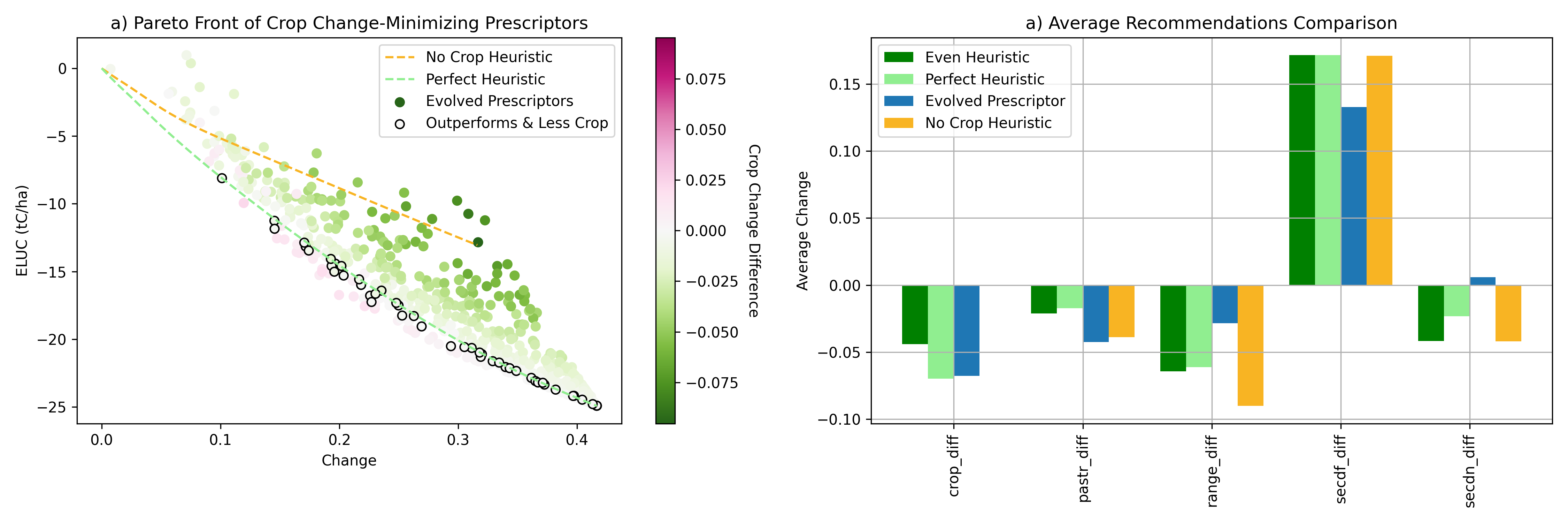}\\[-1ex]
    \caption{Heuristics and Evolved Prescriptors with crop-change minimization as a third objective. ($a$) The final Pareto front of the Evolved Prescriptors is plotted against the Perfect and No Crop Heuristics. Each Evolved Prescriptor is colored according to how different its average prescribed crop change is from the Perfect Heuristic. The points that are outlined identify Evolved Prescriptors that outperform the Perfect Heuristic in all three objectives. ($b$) The average prescriptions of one such Evolved Prescriptor. It reduces pasture more in order to reduce crops less while still outperforming the heuristics along all objectives}
    \label{fg:cropchange}
\end{figure}

Three main findings resulted from this experiment. First, the No Crop heuristic performs quite poorly in the other two objectives, demonstrating that more sophisticated optimization is necessary to incorporate this constraint. Second, evolution was able to find solutions trading off all three objectives, allowing a decision-maker to fix the amount of crop to change before selecting their preferred solution. Third, evolution was able to discover solutions that outperformed the Perfect Heuristic in all three objectives. The trade-off can be seen in Figure \ref{fg:cropchange}$a$. The Pareto front is again plotted as in Figure \ref{fg:heuristics}, but this time, each Evolved Prescriptor is colored by the difference in how much it changed cropland on average compared to the Perfect Heuristic at the same level of change. There is a clear color gradient from more to less crop change perpendicular to the Pareto front such that generally the candidates that prescribed more crop change did the best and those that prescribed less did worse in terms of ELUC and change. Thus, a decision-maker can first choose how much crop they want to change, and then select the best prescriptor at that level along the Pareto front.

The solutions that outperform the Perfect Heuristic in all three objectives are shown in Figure~\ref{fg:cropchange}$a$ outlined in black. Each point represents a prescriptor that achieves a better ELUC with less change than the Perfect Heuristic while also prescribing less crop change. Figure \ref{fg:cropchange}$b$ shows the average prescription of one such point in comparison to the heuristics in order to determine how it did so. The Evolved Prescriptor changed slightly less crop but compensated by removing more pasture. This strategy allowed it to prescribe less crop change, less change overall, and achieve a lower ELUC than the perfect heuristic, showing the power of evolution in the case of three objectives. Ultimately, the results of this experiment show the ability of the proof-of-concept system to incorporate realistic constraints such as those on food production in order to make it practical in real-world decision-making.

\subsection{Incorporating Human Expertise}
\label{sc:rhea}

\begin{figure*}[!t]
    \centering
    \includegraphics[width=\linewidth]{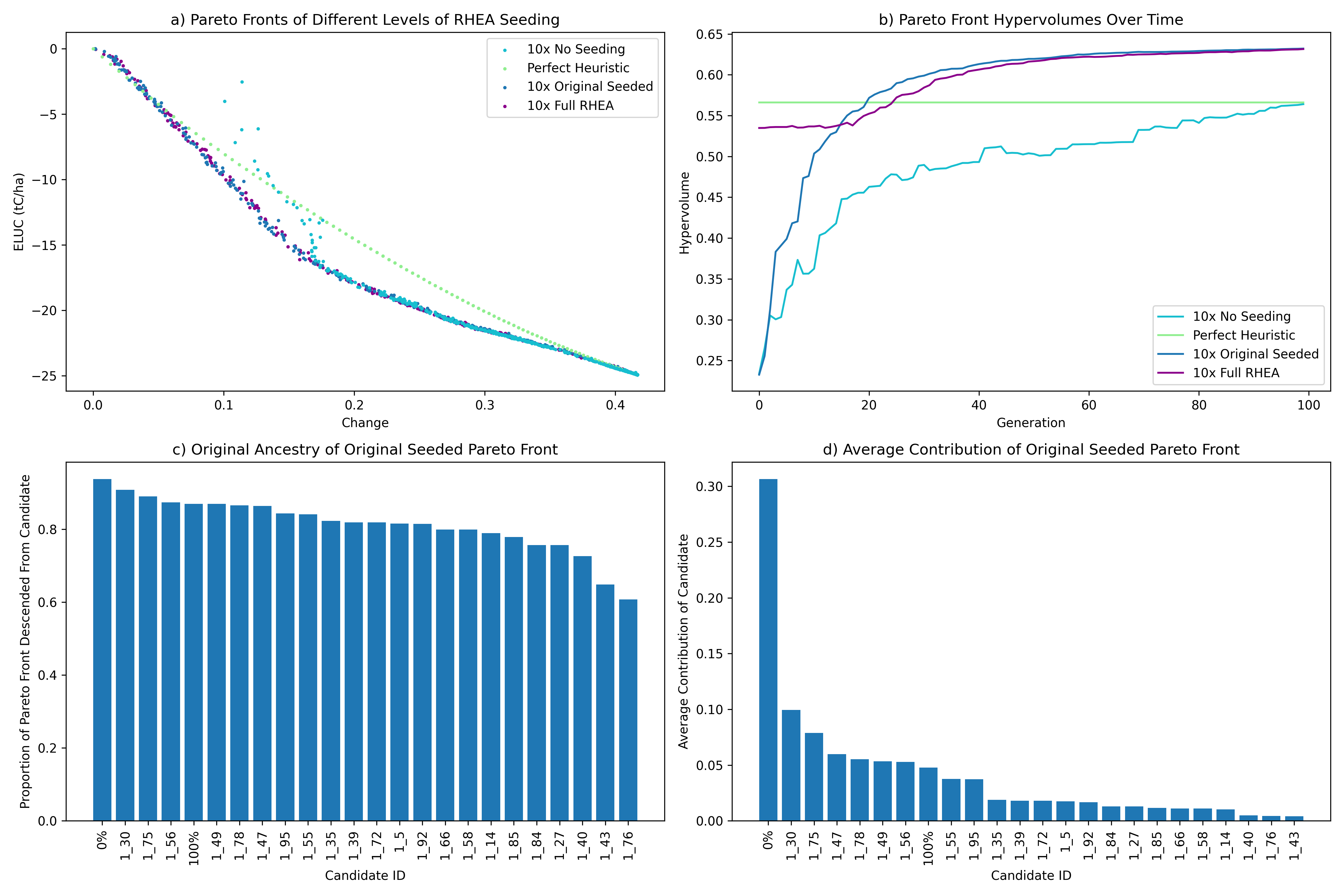}\\[-1ex]
    \caption{The value of incorporating human expertise. ($a$) Combined Pareto fronts over 10 trials for: evolution run without initial seeding, the Perfect Heuristic, the original seeded Evolved Prescriptors (evolved with two extreme seeds), and a full RHEA evolution (with an extended seeding of the population) evaluated on the test dataset. The two extreme seeds achieved as strong results as extended seeding. However, without any seeds evolution fails to find the top half of the Pareto front. ($b$) The hypervolume comparison visualizes these results clearly as well. ($c$) The proportion of candidates in the Evolved Prescriptors Pareto front from Section \ref{prescriptions-section} that have a given candidate from the first generation in its ancestry. The distilled 0\% and 100\% heuristics are ancestors of 94\% and 87\% of the Pareto front, respectively. ($d$) The average contribution of each first-generation candidate to the same Pareto front. The distilled 0\% perfect heuristic makes up on average over 30\% of total ancestry for the candidates in the final Pareto front. Thus, even limited seeding can have a significant effect on the quality of solutions, suggesting that RHEA is instrumental in constructing practical solutions to complex decision-making problems like land-use optimization.}
    \label{fg:rhea}
\end{figure*}

When developing solutions to global societal problems with AI, incorporating expert human knowledge may improve performance while also adding accountability and explainability \cite{meyerson:neurips24}. The first step was taken in Section \ref{prescriptions-section}, where two hand-designed solutions were included in the initial prescriptor population through RHEA. 

In more detail, RHEA consists of (1) defining the problem so that experts can develop solutions to it, (2) gathering a set of solutions from the experts, (3) distilling these solutions into a population of evolvable models such as neural networks, and (4) discovering better models by evolving this population. The land-use optimization problem was already defined with context, action, and outcome variables and was therefore compatible with the framework. Next, the behaviors for prescribing no change and all change to secondary forest were identified as useful extremes. In order to distill these behaviors, two sets of labels for 10\% of the training dataset were created in order to run backpropagation. One was constructed by not changing the land use at all and the other changed as much land as possible to secondary forest. Then the AdamW optimization algorithm was run to minimize mean squared error over 15 epochs to targets that prescribed either no change or full change for a sampling of land-use cases. The resulting two trained prescriptors were then injected into the initial population and evolution was run.

In order to determine how useful it is in general to incorporate human expertise in this manner, two further experiments were run. One was an ablation study where no RHEA was performed and the entire initial population was randomly orthogonally initialized. The second one was an extended seeding study where the Perfect Heuristic was queried with thresholds from 5\% to 95\% in increments of 5\%, prescriptors were trained with backpropagation to imitate the results of each query, and the resulting 19 seed models were added to the initial population together with the original two extreme prescriptors. The original and the new experiments were each run 10 times. Pareto fronts for each generation were created by combining all 10 trials' candidates and then forming the Pareto front of the combined population. These Pareto fronts were also used to compute the hypervolume at each generation.

The results, shown in Figure \ref{fg:rhea}$a$, demonstrate the power of incorporating human expertise. When the distilled seed models were removed, even with 10 trials, evolution was unable to discover the top half of the Pareto front, missing out on the most realistic decision-making use cases (i.e.\ those with small changes). In Figure \ref{fg:rhea}$b$, the hypervolume of the No Seeding experiment barely reaches that of the Perfect Heuristic. However, just including one seed on each end of the Pareto front, as in the original seeded experiment, allowed evolution to achieve better results than the Perfect Heuristic. Additional seeds in the full RHEA case did not significantly improve the final solutions. Thus, simply injecting two extreme seeds based on human expertise was sufficient for this task.

The results of RHEA in a single run of the original two-seed experiment in Section \ref{prescriptions-section} are further illustrated in Figures \ref{fg:rhea}$c$ and \ref{fg:rhea}$d$ (the results with other trials were qualitatively similar). Figure \ref{fg:rhea}$c$ shows the proportion of the final Pareto front that has a given candidate from the first generation as an ancestor. The no-change and full-change seed behaviors are ancestors of 94\% and 87\% of the final Pareto front, thus contributing to almost every prescriptor. The magnitude of that contribution is measured in Figure \ref{fg:rhea}$d$ by constructing a tree of parents for each prescriptor in the final Pareto Front and counting how many times each first-generation candidate was included in it. On average a model in the Pareto front is made up of over 30\% of the distilled no-change behavior, mutated and crossed-over throughout the generations of evolution. This result aligns with the fact that the No Seeding experiment was unable to find the top left corner of the Pareto front: The seeded experiment needed to use the no-change solution frequently to fill in this top half. Very little behavior is taken from other randomly initialized models, showing the importance of the seeded model. Distilling this particular behavior was crucial in order to achieve the final Pareto front. The ultimate performance increase and explainability show how incorporating just a small amount of human feedback can improve the automatic decision-making process. In the future, as more real-world factors are taken into account, human expertise is likely to become even more valuable, and can be further applied with RHEA.

\subsection{Interactive Evaluation}

An interactive demo of the trained/evolved models is provided at
\url{https://landuse.evolution.ml}. It allows the user to explore different locations,
observe the prescribed actions for them, and see their outcomes. It is
also possible to modify those actions and see their effect, thus
evaluating the user's skills as a decision-maker compared to that of
machine learning.

\begin{figure*}[t]
    \centering
    \includegraphics[width=\linewidth]{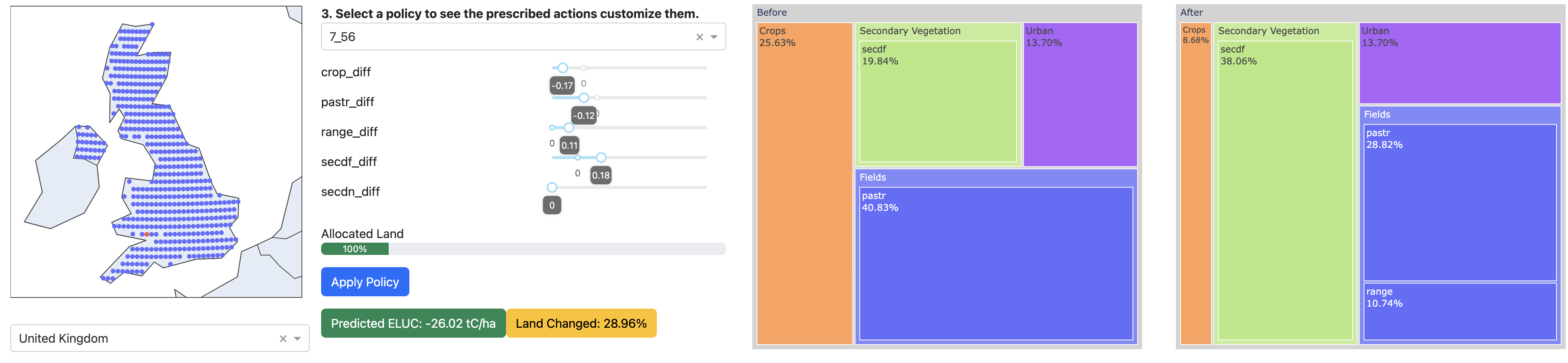}\\[-1ex]
    \caption{A suggested land-use change for a given location (screenshots from the demo at \url{https://landuse.evolution.ml}). The location is indicated by the red dot among the UK grid cells. One Evolved Prescriptor is chosen from the middle region of the Pareto front spanning minimal change and minimal ELUC.  The current land use is shown on the left chart and the recommended one on the right chart, as well as the sliders on the left. This prescriptor recommends decreasing pasture and crops and increasing range and secondary forest, resulting in a 26.02 tC/ha decrease in carbon emissions with a 28.96\% land-use change. The user can then select different solutions from the Pareto front and modify the sliders manually to explore alternatives
    }
    \label{fg:demo-uk}
\end{figure*}

As an example, Figure~\ref{fg:demo-uk} shows the grid cells from which
the user has selected one. The system recommends a change in land use
that results in a 26.02 tC/ha reduction in carbon emissions with a 28.96\% land-use change.
Using the sliders, the user can then explore alternative tradeoffs and
modify specific elements of each suggestion. For each variation, the user
can then obtain the estimated carbon reduction and land-use change percentage, and
in this manner gain an understanding of the possible attainable solutions
and confidence in the preferred choice.

\section{Future Work}
\label{sc:future}

The project so far has resulted in a promising proof-of-concept for utilizing ESP in land-use planning.  In order to make it usable for real-world decision-makers, it needs to be extended with more fine-grained data. The models so far did not distinguish between the different types of crops and only partly considered regional variations in how land-use changes affect carbon stocks in soil and vegetation (through training different models for different regions). The assumed values for carbon stocks in soil and vegetation represent conditions of the recent past \citep{hansis:gbc15}; they do not consider environmental effects such as disturbances due to fires or droughts or better vegetation growth due to CO$_2$ effects or a warmer climate, which can substantially impact terrestrial CO$_2$ fluxes \citep[e.g.][]{aragao:natcom18,pongratz:currclimcr21,rosan:commee24}. Such additional features could be added as context input to the predictor and prescriptor models, and the actions could be updated accordingly to prescribe more detailed policies. Further, estimates of CO$_2$ fluxes from land-use change also depend on the spatial resolution of the land-use change dataset \citep{ganzenmuller:erl22,bulton:natcom22}. Extending the project with such details should lead to more precise and more fine-scale estimates of the impact of land-use change decisions, allowing decision-makers to utilize these dimensions in the real world.

Another immediate future work direction consists of improving the predictor accuracy through ensembling. Given that models trained on specific regions perform the best on those regions, there is a good chance that an ensemble of them will do even better. Moreover, there may be a benefit of including LinReg, RF, and NeuralNet versions of these models in the ensemble as well. There are several ensembling techniques in the Project Resilience codebase, such as averaging, sampling, and confidence-based ensembling, and contributors are encouraged to submit more in the future. The effect of such alternatives will be evaluated in future work, both on prediction accuracy and on driving the search for better prescriptors. The hypervolume of the Pareto front can be used to compare the different approaches quantitatively.

The predictor creates point estimates of ELUC. A technique called
Residual Input-Output kernel \citep[RIO;][]{qiu:iclr20} can be used to estimate
confidence in these predictions. The method builds a Gaussian Process
model of the residual error, quantifying the uncertainty. RIO is part
of the Project Resilience codebase but was not yet used in the study in
this paper.

Project Resilience also provides an alternative approach to
prescription: Representing the prescriptor as a set of rules
\citep{shahrzad:arxiv22}. This approach has the advantage that the behavior is
transparent and explainable, as demonstrated in prior experiments in
other tasks (including NPI optimization \citep{miikkulainen:ieeetec21}). Rule-set evolution on
land-use optimization will be evaluated in the future.

To make the system more immediately useful to decision-makers, the cost metric could be refined to take into account region-specific costs of changing each particular land-use type to another.
The approach can also be extended with preferences and further objectives. For instance, the decision maker could choose which crops should be preferred, which may be more actionable than the large-scale changes in the current dataset. It may be possible to optimize additional objectives such as minimizing nitrate pollution and water consumption, and maintaining or increasing food production, similar to how minimizing crop change was optimized in Section \ref{sc:crop}. Further, instead
of planning for a single year, it might be possible to develop prescriptors that
recommend land-use changes for the next several years: For instance, what could
be done in the next five years to reduce ELUC by a given target percentage. 
As data and simulations become more accurate in the future, such refined 
and extended decisions should become possible.

Even though land-use change is an important factor, it alone is not enough to reduce emissions to net zero. However, the framework designed in this project can be applied to other decision-making use cases as long as available context, action, and outcome data exist. A comprehensive such effort, covering energy policy and other factors, may eventually help reach climate-change goals.

\vspace*{-2ex}
\section{Conclusion}

Land-use policy is an area relevant to climate change where local decision-makers
can have a large impact. In this paper, historical data and simulation technology
are brought together to build an efficient machine-learning model that can
predict the outcomes of these decisions efficiently for different contexts.
An evolutionary optimization process is then used to identify effective
policies, balancing the cost of land-use change and its effect in reducing
carbon emissions. Machine learning methods can thus play an important role in
empowering decision-makers to act on climate change issues.

\begin{appendix}
\section{Appendix: Using Location as a Proxy for Climate Information}\label{appendixA}
\begin{figure*}[!t]
    \includegraphics[width=\linewidth]{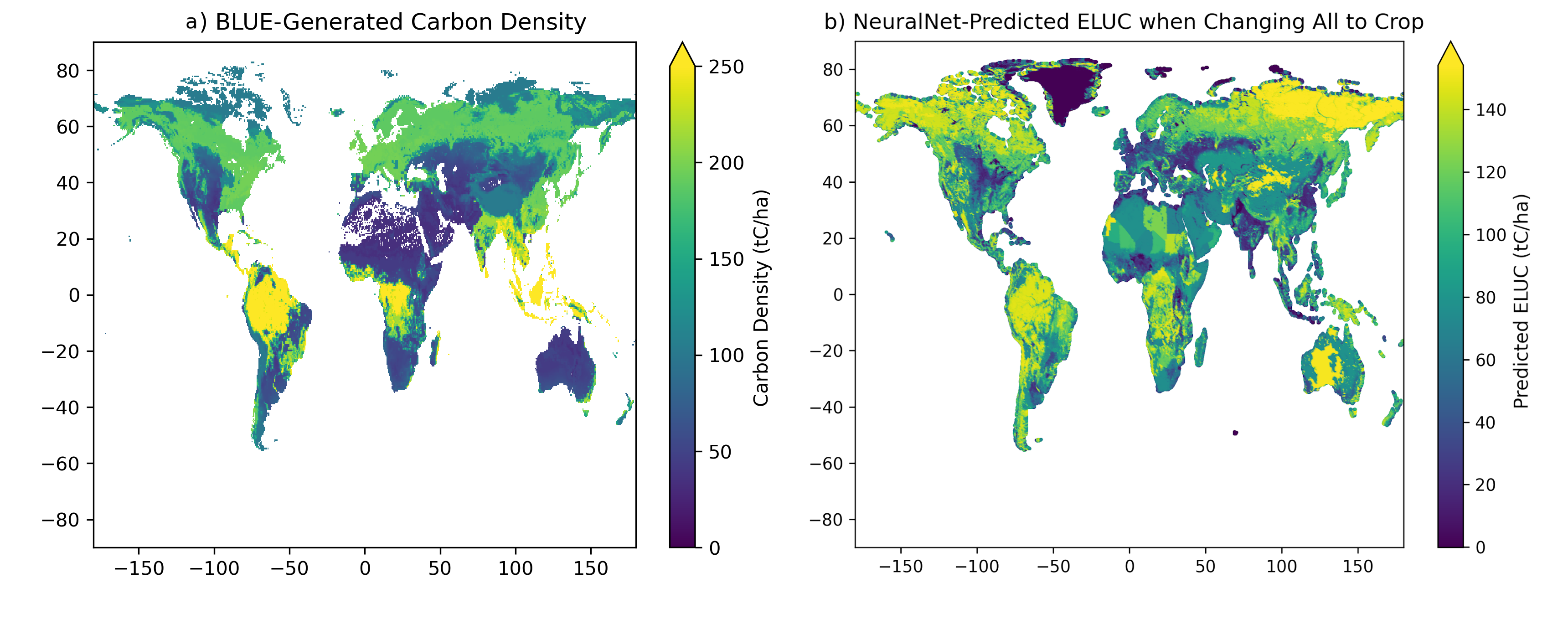}\\[-1ex]
    \caption{Location as a Proxy for Climate. Different climates have different plant functional types (PFTs), which in turn have different carbon densities. ($a$) The carbon density, or potential carbon release, calculated with the BLUE model across the globe. ($b$) NeuralNet-predicted ELUC when each land cell is converted entirely to crop, acting as a proxy for carbon density. The same general qualitative trends appear across the globe with some outliers, showing the ability of the NeuralNet to model climate information implicitly through plant functional types.}
    \label{fg:latlon}
\end{figure*}

Different locations have different emissions from land-use change due to the climate. They are affected by both the initial makeup of the land use and the conditions of the cell such as weather, soil, and type of vegetation. The BLUE model takes this type of vegetation into account as plant functional types (PFTs) when calculating ELUC. In order to test if the predictor model encodes such climate information, an experiment was run to see if similar trends arose in both the predicted ELUC from the NeuralNet and the carbon densities (i.e.\ potential for carbon release) computed from the BLUE model using plant functional types. Similarities in the results would indicate that the predictor model is able to learn climate information from plant functional types even when they are not explicitly passed in as features.

The carbon densities from the BLUE model are shown in Figure \ref{fg:latlon}$a$. To compute the corresponding values in the model, first note that crops have a significantly lower carbon density than vegetation (this effect can be seen e.g.\ in Figure \ref{fg:all-model-heatmaps}, where conversion to crop from anything else results in a high ELUC). Thus, converting all land to crop would be similar to releasing all the carbon from the land, and the resulting ELUC can thus be seen as an estimate of its carbon density. To obtain this estimate, all land types were converted to crops globally across all data points in the year 2020, and the resulting ELUC predicted for this change using the NeuralNet. The result is plotted on a map in Figure \ref{fg:latlon}$b$.

Similar qualitative trends arise in both plots, with core primary forest regions like the Amazon and the Congo Basin having high carbon densities and predicted ELUCs. Secondary forest regions such as the eastern US, southern Canada, and southern China were picked up by the predictor as having a more moderate carbon density/ELUC. Additionally, drier regions like Central Asia tended towards lower carbon densities and ELUC values. The scales do not perfectly align, however it is more important to view overall trends rather than numeric similarities.

However, qualitatively higher ELUC values than carbon densities appeared in the Great Victoria Desert, northeastern Siberia, and northern Africa. This result was due to the fact that LUH2 classifies these regions as almost entirely primary nonforest vegetation, which the NeuralNet erroneously predicted to release a large amount of carbon. LUH2 uses country-level Food and Agriculture Organization (FAO) statistics when there is not enough data, and therefore defaults to a uniform distribution. While the predictor was able to pick up on trends in the other types of vegetation, it overestimated the ELUC for such regions.

To conclude, similar trends arise overall in carbon density estimates in BLUE and the model, indicating that the model captures climate information even when it is not explicitly given to it as features. Further alignment could be performed in the future using more specific data, however for the purposes of this proof-of-concept study, the amount of information encoded in the current features is sufficient.

\end{appendix}

\vspace*{2ex}
\paragraph{Acknowledgments}
We would like to thank the BLUE, LUH2, and Project Resilience teams, in particular Amir Banifatemi, Prem Krishnamurthy, Gyu Myoung Lee, Gillian Makamara, Michael O’Sullivan, Julia Pongratz, and Jennifer Stave.

\paragraph{Funding Statement}
This work received no specific grant from any funding agency, commercial or not-for-profit sectors.

\paragraph{Competing Interests}
The authors report no competing interests.

\paragraph{Code and Data Availability Statement}
Code and data for the experiments are available at \url{https://github.com/Project-Resilience/mvp/tree/main/use_cases/eluc}. Permanent snapshots of the data and code are archived on Zenodo at: \href{https://zenodo.org/records/15226287}{10.5281/zenodo.15226287} and \href{https://zenodo.org/records/15231467}{10.5281/zenodo.15231467}


\paragraph{Author Contributions}
Conceptualization: RM, OF, EM, JB, HC, BH;
Data Curation: CS, OF, DY, JB, HC;
Formal Analysis: EM, OF, DY;
Investigation: RM, OF, DY, EM, BH;
Methodology: RM, OF, DY, EM, BH;
Project Administration: RM, OF, BH;
Software: OF, EM, DY;
Supervision: RM, OF, BH;
Validation: RM, OF, EM, DY, BH;
Visualization: RM, OF, EM, DY;
Writing – Original Draft: RM, OF, DY, BH;
Writing – Review \& Editing: RM, OF, DY, EM, CS, BH.

\bibliographystyle{plain}
\bibliography{main}

\end{document}